\documentclass[sigconf]{acmart}

\AtBeginDocument{%
  }

\copyrightyear{2026}
\acmYear{2026}
\setcopyright{cc}
\setcctype{by}
\acmConference[SIGIR '26]{Proceedings of the 49th International ACM SIGIR Conference on Research and Development in Information Retrieval}{July 20--24, 2026}{Melbourne, VIC, Australia}
\acmBooktitle{Proceedings of the 49th International ACM SIGIR Conference on Research and Development in Information Retrieval (SIGIR '26), July 20--24, 2026, Melbourne, VIC, Australia}
\acmDOI{10.1145/3805712.3809639}
\acmISBN{979-8-4007-2599-9/2026/07}

\acmConference[SIGIR'26]{}{July 20--24,
  2026}{Melbourne, Australia}

\usepackage{algorithm}
\usepackage{algorithmic}
\usepackage{multirow}
\usepackage{booktabs}
\usepackage{xcolor}
\usepackage{amsmath}
\usepackage{pifont}
\usepackage{tabularx}

\begin{document}

\title{Denoise and Align: Diffusion-Driven Foreground Knowledge Prompting for Open-Vocabulary Temporal Action Detection}

\author{Sa Zhu}
\affiliation{%
  \institution{Institute of Information Engineering, \\ Chinese Academy of Sciences\\ School of Cyber Security, University \\  of Chinese Academy of Sciences \\ State Key Laboratory of Cyberspace Security Defense}
  \city{Beijing}
  \country{China}}
\email{zhusa@iie.ac.cn}

\author{Wanqian Zhang}
\authornote{Corresponding author.}
\affiliation{%
  \institution{Institute of Information Engineering, \\ Chinese Academy of Sciences\\ }
  \city{Beijing}
  \country{China}}
\email{zhangwanqian@iie.ac.cn}

\author{Lin Wang}
\affiliation{%
  \institution{Hangzhou Dianzi University }
  \city{Hangzhou}
  \country{China}}
\email{wanglin@hdu.edu.cn}

\author{Jinchao Zhang}
\affiliation{%
  \institution{Institute of Information Engineering, \\ Chinese Academy of Sciences\\State Key Laboratory of Cyberspace Security Defense }
  \city{Beijing}
  \country{China}}
\email{zhangjinchao@iie.ac.cn}

\author{Cong Wang}
\affiliation{%
  \institution{Control Science and Engineering, \\ Zhejiang University }
  \city{Hangzhou}
  \country{China}}
\email{cwang85@zju.edu.cn}

\author{Bo Li}
\affiliation{%
  \institution{Institute of Information Engineering, \\ Chinese Academy of Sciences\\ State Key Laboratory of Cyberspace Security Defense }
  \city{Beijing}
  \country{China}}
\email{libo@iie.ac.cn}

\renewcommand{\shortauthors}{Sa Zhu et al.}

\begin{abstract}
Open-Vocabulary Temporal Action Detection (OV-TAD) aims to localize and classify action segments of unseen categories in untrimmed videos, where effective alignment between action semantics and video representations is critical for accurate detection. 
However, existing methods struggle to mitigate the semantic imbalance between concise, abstract action labels and rich, complex video contents, inevitably introducing semantic noise and misleading cross-modal alignment.
To address this challenge, we propose DFAlign, the first framework that leverages diffusion-based denoising to generate foreground knowledge for the guidance of action–video alignment. 
Following the \textit{`conditioning, denoising and aligning'} manner, we first introduce the Semantic-Unify Conditioning (SUC) module, which unifies action-shared and action-specific semantics as conditions for diffusion denoising.
Then, the Background-Suppress Denoising (BSD) module generates foreground knowledge by progressively removing background redundancy from videos through denoising process. 
This foreground knowledge serves as effective intermediate semantic anchor between video and text representations, mitigating the semantic gap and enhancing the discriminability of action-relevant segments.
Furthermore, we introduce the Foreground-Prompt Alignment (FPA) module to inject extracted foreground knowledge as prompt tokens into text representations, guiding model's attention towards action-relevant segments and enabling precise cross-modal alignment.
Extensive experiments demonstrate that our method achieves state-of-the-art performance on two OV-TAD benchmarks. The anonymous code repository is provided as follows: \url{https://anonymous.4open.science/r/Code-2114/}.

\end{abstract}



\maketitle

\section{Introduction}
Temporal action detection (TAD) aims to localize and classify action instances within untrimmed videos, which is a fundamental task in video understanding~\cite{zhu2025uneven, fang2023uatvr} and has been explored by many works~\cite{chen2025temporal, pang2025context, kumar2025stable, kim2025prediction, yoshida2025action}.
However, conventional TAD methods rely heavily on large-scale annotated data for supervised training, the collection of such data is labor-intensive and time-consuming, limiting their practicality in real-world scenarios.

\begin{figure}[!t]
  \centering
  \includegraphics[width=1.0\linewidth]{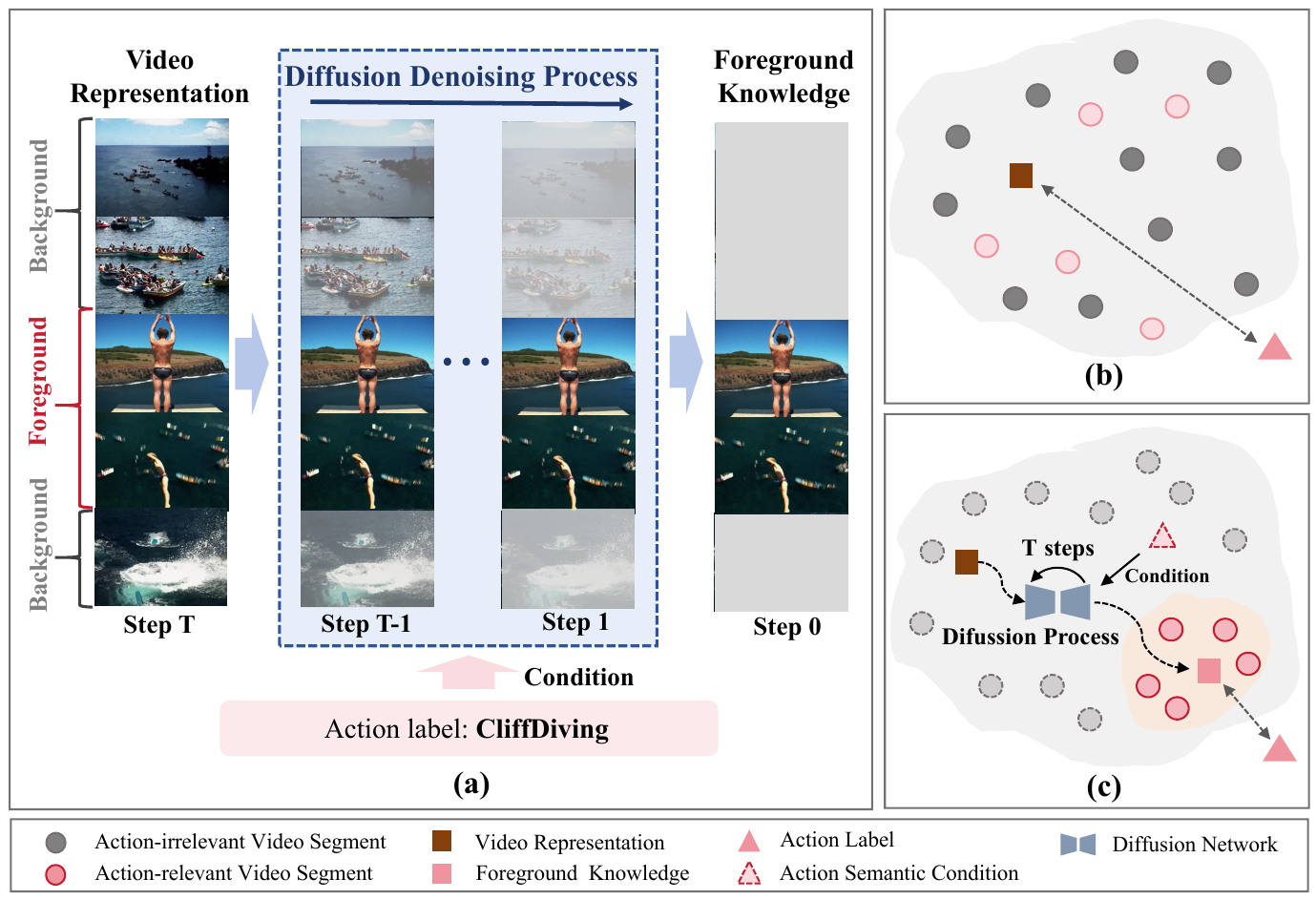}
  \caption{
  (a) The diffusion model could progressively suppress background information in video representations guided by the action semantic condition, generating foreground knowledge.
  We illustrate the visualizations of video segment embeddings and action label embeddings produced by (b) baseline Ti-FAD and our method (c).
  The diffusion-generated foreground knowledge acts as an intermediate semantic anchor between video and text representations, promoting tighter clustering of action-relevant segments and effectively narrowing the semantic distance between action labels and action-relevant video segments.
  }
  \label{Motivation}
\end{figure}

To address this challenge, Open-Vocabulary Temporal Action Detection (OV-TAD), also known as Zero-Shot Temporal Action Detection (ZSTAD), has been proposed to detect action categories that are unseen during training. 
Due to the absence of target action labels in the training phase, existing OV-TAD methods predominantly align video representations with text-based action descriptions derived from pre-trained vision–language models (VLMs) for action detection. 
Consequentially, effectively aligning the action semantics with video features is crucial for unseen action detection. 
Early OV-TAD approaches~\cite{nag2022zero, ju2022prompting, phan2024zeetad} enhance cross-modal alignment by introducing learnable textual tokens within the text encoder. 
Recently, Ti-FAD~\cite{lee2024text} introduces the cross-attention to infuse textual information into visual features, while CSP~\cite{wangconcept} proposes the concept-guided semantic projection framework to project video and action features into a unified concept space. 

Despite their encouraging performance, these methods largely overlook the fundamental issue of semantic imbalance between action labels and video contents. 
Specifically, action labels are typically short and abstract, providing limited semantic coverage. 
On the contrary, untrimmed videos contain abundant background and context-irrelevant information, inevitably introducing noise and misleading cross-modal alignment.
As a result, explicitly extracting discriminative foreground knowledge from video representations, while suppressing background redundancy, becomes crucial for bridging semantic gap during alignment and improving generalization of OV-TAD models on unseen action categories.

Recently, Diffusion models (DMs) have demonstrated superior conditional generations by progressively removing redundant noise from input features via conditioning-guided denoising process~\cite{li2023momentdiff, nag2023difftad, wang2024diffusion}. 
Motivated by this, we investigate to employ a diffusion-based denoiser that incrementally suppresses action-irrelevant background information in video representations. 
As illustrated in Figure~\ref{Motivation} (a), this iterative denoising process yields discriminative foreground knowledge by suppressing background redundancy. 
Furthermore, as in Figure~\ref{Motivation} (b) and (c), as the intermediate semantic anchor between video and text representations, the extracted foreground knowledge effectively clusters action-relevant segments and mitigates the semantic gap with action labels, thereby enabling more precise cross-modal alignment.

In this paper, we propose a diffusion-based open-vocabulary temporal action detection framework, termed as DFAlign. 
The framework consists of three key components: conditioning, denoising and aligning. 
First, we introduce a Semantic-Unify Conditioning (SUC) module to construct action semantic conditions for diffusion denoising. 
We argue that conditioning on each action label separately and repeatedly not only incurs high computational cost, but also overlooks shared information across different actions. 
Instead, SAC elicits action-shared semantics via large language model (LLM) prompting, which is further unified with action-specific semantics through a Foreground Semantic Aggregator. 
This design not only adaptively integrates action-shared semantics into conditioning, but also avoids the computational overhead of repeated diffusion processes.
Next, we propose Background-Suppress Denoising (BSD) module to generate foreground knowledge. 
Guided by unified action semantic conditions, the action-irrelevant segment information is progressively suppressed from video contents via diffusion denoising process. 
The generated foreground knowledge could serve as an effective intermediate semantic anchor between video and text representations, reducing the semantic gap and enhancing the discriminability of action-relevant segments.
Finally, we propose the Foreground-Prompt Alignment (FPA) module that leverages the extracted foreground knowledge as explicit guidance for action–video alignment. 
By injecting foreground knowledge as prompt tokens into text representations, FPA interact with video features, guiding the model’s attention toward action-relevant segments and enabling more precise alignment with action semantics, thereby improving detection performance.

Our contributions can be summarized as follows:
\begin{itemize}
  \item We propose DFAlign, a novel diffusion-based framework for OV-TAD, which employs a background diffusion denoiser to transform video representations into foreground knowledge under the guidance of action semantics. This foreground knowledge could guide the learning of discriminate action-relevant segments and reducing the semantic gap, facilitating more precise cross-modal alignment.

  \item We design a Semantic-Unify Conditioning (SUC) module, which unifies action-shared and action-specific semantics to construct conditions for diffusion denoising. In addition, we propose a Foreground-Prompt Alignment (FPA) module that injects the extracted foreground knowledge as prompt tokens into text representations to align with video features, effectively alleviating the semantic gap.
  
  \item Comprehensive experimental results validate the effectiveness of the proposed method, achieving superior OV-TAD performance that surpasses state-of-the-art methods.
\end{itemize}

\section{Related Work}

\textbf{Temporal Action Detection (TAD)} ~\cite{kim2024te, liu2024end, zhu2024dual, jiang2024exploring, singh2024semi, denize2024comedian, zeng2024unimd, chen2025temporal, kim2025digit, pang2025context} aims to classify and localize action instances within a closed set of predefined action categories.
Early approaches typically adopt a two-stage pipeline, where candidate temporal proposals are first generated and then refined through action classification and boundary adjustment.
Representative methods include G-tad~\cite{xu2020g}, RTD-Net~\cite{tan2021relaxed} and ATAG~\cite{chang2022augmented}. 
More recently, transformer-based methods have been proposed to streamline this process by unifying proposal generation and classification into a single-stage framework. 
For instance, 
TriDet~\cite{shi2023tridet} introduces triplet point modeling for more accurate boundary localization. 
DyFADet~\cite{yang2024dyfadet} dynamically aggregates multi-scale features to better adapt to varying action durations, 
while DiGIT~\cite{kim2025digit} introduces a multi-dilated gated encoder and a central–adjacent region integrated decoder to capture features across diverse receptive fields and provide each detection query with a more comprehensive contextual view.
Despite the remarkable advancements, most TAD methods remain constrained to closed-set scenarios, limiting their applicability in open-world settings where unseen actions may emerge.

\vspace{0.15cm}

\noindent
\textbf{Open Vocabulary Temporal Action Detection (OV-TAD)} aims to localize and classify actions unseen during training~\cite{zhu2026decompose}. 
This field commonly leverages the generalization ability of pre-trained vision–language models such as CLIP, classifying unseen action proposals by performing visual–textual alignment.
The pioneering work, Efficient-Prompt~\cite{ju2022prompting}, introduces a prompt-based two-stage framework where activity proposals are detected via alignment between proposal features and learnable token-prompted action labels. 
However, its two-stage design suffers from interference between localization and classification. 
To alleviate this issue, STALE~\cite{nag2022zero} reduces error propagation across stages, DeTAL~\cite{li2024detal} decouples proposal generation and classification through separate networks, and ZEETAD~\cite{phan2024zeetad} incorporates a dual-localization module with semantic masking and lightweight CLIP-based adapters for detection.
More recently, Ti-FAD~\cite{lee2024text} introduces a cross-attention to infuse textual information into visual features, enhancing visual-textual interactions for action detection. 
In parallel, CSP~\cite{wangconcept} projects video features into a semantic concept space to enhance the semantic consistency of learned action representations. 

Despite notable progress, most methods overlook the semantic imbalance between action labels and video content: action labels are typically too concise to reflect the rich semantics of action proposals, while untrimmed videos contain substantial background and context-irrelevant information that introduces noise and hampers cross-modal alignment.
In contrast, we propose DFAlign, which leverages diffusion models to extract foreground knowledge from videos. This foreground knowledge guides the model toward precise alignment with action-relevant segments and effectively reduces the semantic gap, thereby enhancing unseen action detection.

\vspace{0.15cm}

\noindent
\textbf{Diffusion Models} ~\cite{sohl2015deep, ho2020denoising} are generative models based on stochastic diffusion processes, which progressively add noise to data samples and learn to reverse this process via iterative denoising. 
Developments in diffusion models have achieved great achievements in vision generation tasks~\cite{epstein2023diffusion, zhu2023conditional, yue2024efficient, li2025dual, cai2025diffusion}, showing impressive fidelity and controllability in image synthesis.
Recently, motivated by their strong ability to incorporate semantic guidance during the denoising process, diffusion models have recently been extended beyond image generation to various video-related tasks. 
For text–video alignment, DITS~\cite{wang2024diffusion} employs diffusion to bridge the modality gap by denoising a text-conditioned latent toward the corresponding video representation.
For video moment localization, MomentDiff~\cite{li2023momentdiff} and CVDN~\cite{liu2023conditional} leverage diffusion to regress temporal boundaries, progressively refining random noise into accurate moment spans under multimodal guidance.
In temporal action detection task, ADI-Diff~\cite{foo2024action} treats the starting point, ending point and action-class outputs as three AD images, and reformulates action detection as an AD-image generation task through diffusion process, while DiffTAD~\cite{nag2023difftad} diffuses action proposals into noise and conditionally denoises them to recover precise action spans.

However, cross-modal generation incurs significant computational costs, as the process typically begins from random Gaussian noise, and obtaining spans from noise fails to tap into the DM’s content generation capabilities.
In contrast, our method initializes the diffusion process from video representations, and progressively denoises it to remove action-irrelevant background, producing foreground knowledge essential for alignment. 
This design not only reduces computational overhead, but also narrows the semantic gap and supports effective generalization to unseen actions.

\section{Method}

\subsection{Problem Formulation}
Given a training set of untrimmed videos $D_{train}$, each video is represented as a sequence of visual features $X=\{x_i\}_{i=1}^{N}$, where $N$ denotes the number of segments (a few sequences of frames).  
The corresponding annotations are defined as $Y=\{s_m, e_m, c_m\}^{M}_{m=1}$, where $s_m$ and $e_m$ indicate the start and end timestamps of the $m$-th action, and $c_m \in \mathcal{C}_{train}$ denotes the action categories for training.
In the open-vocabulary setting, the training and testing label sets are disjoint, i.e., $\mathcal{C}_{train} \cap \mathcal{C}_{test} = \emptyset$.
The goal of Open-Vocabulary Temporal Action Detection (OV-TAD) is to identify actions from unseen categories and predict the start and end timestamps of their corresponding temporal segments within the test videos. 

\subsection{Overall Framework}
\begin{figure*}[!t]
\centering
  \includegraphics[width=0.97\textwidth]{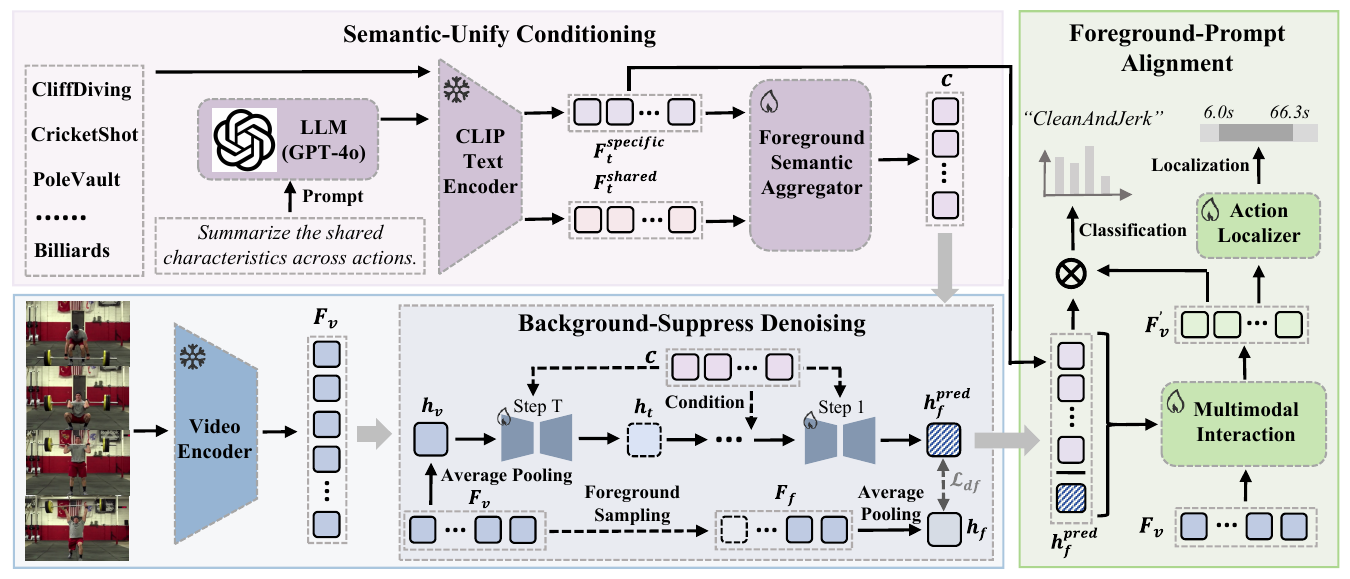}
  \caption{
  Overview of the proposed framework, which consists of three key modules: Semantic-Unify Conditioning (SUC), Background-Suppress Denoising (BSD), and Foreground-Prompt Alignment (FPA).
  The SUC module constructs diffusion conditions by jointly modeling action-shared and action-specific semantics. Guided by these conditions, the BSD module progressively suppresses background information in video representations via diffusion denoising, yielding foreground knowledge. Finally, the FPA module injects the foreground knowledge as prompt tokens to enable cross-modal interaction for unseen action detection.
  }
 \label{framework}
\end{figure*}

Figure~\ref{framework} illustrates the overall architecture of our proposed framework. Video and action representations are first extracted by pretrained visual and textual backbones. The Semantic-Unify Conditioning (SUC) module then constructs action semantic conditions by integrating action-shared and action-specific semantics. Guided by these conditions, the Background-Suppress Denoising (BSD) module progressively removes background redundancy from video representations through a diffusion denoising process, producing discriminative and informative foreground knowledge. Finally, the Foreground-Prompt Alignment (FPA) module injects the extracted foreground knowledge into text representations to facilitate cross-modal interaction, thereby reducing the semantic gap and enabling more precise action classification and localization.

\subsection{Feature Extraction}
Following prior works~\cite{lee2024text, wangconcept}, we first extract visual features using a visual encoder $\Phi_{vis}(\cdot)$. 
The resulting features are processed by masked convolutional layers to model temporal dynamics, producing the final video representation ${F}_{v} \in \mathbb{R}^{N_v \times D}$, where $N_v$ denotes the number of temporal segments and $D$ is the feature dimension. 
Similarly, for textual representation, we adopt the class names of actions as text inputs and extract text features via CLIP’s text encoder $\Phi_{txt}(\cdot)$. 
These features are then projected into a shared embedding space, yielding the final text representation ${F}_{t}^{specific} \subseteq \mathbb{R}^{C \times D}$, where $C$ denotes the number of action classes.

\subsection{Semantic-Unify Conditioning}
The key of our method lies in accurately transforming video representations into foreground knowledge. 
To this end, action semantics should be incorporated to guide the denoising network to progressively suppress background information while preserving action-relevant cues.
A naïve solution is to independently condition the diffusion denoising process on each action label to obtain action-specific foreground representations and then aggregate them. 
However, this strategy suffers from two fundamental limitations. 
First, it incurs prohibitive computational overhead, as the diffusion process must be repeated for each action category. 
Second, such isolated denoising fails to capture semantics shared across actions, thereby limiting generalization to unseen categories.

To overcome these issues, we propose to unify action-shared and action-specific semantics for condition generation.
Specifically, we first provide a large language model (LLM), such as GPT-4o~\cite{hurst2024gpt}, with the set of action categories and prompt it to summarize their shared characteristics, focusing on common motion patterns and body dynamics. 
This process captures action-shared semantics across categories. 
An example prompt is: \textit{"You are given the following action categories $\langle Actions \rangle$: Please summarize the shared characteristics across these action categories, from the perspective of motion patterns and body dynamics." }. 
The resulting textual description is encoded by the CLIP text encoder $\Phi_{txt}(\cdot)$ to obtain the action-shared text representation ${F}_{t}^{shared} \in \mathbb{R}^{D}$, where $D$ denotes the feature dimension.
Then, we concatenate the action-shared text representation ${F}_{t}^{shared}$ with action-specific text representation $F_{t}^{sepcific}$ to form a composite text sequence:
\begin{equation}
F_t = \{{F}_{t}^{shared}; {F}_{t}^{generic} \} \subseteq \mathbb{R}^{(C + 1) \times D}.
\end{equation}

This sequence is fed into the Foreground Semantic Aggregator (FSA) implemented by multi-layer Transformer, which models inter-action relationships and effectively fuses action-shared and action-specific representations into a unified action semantic condition:
\begin{equation}
c = \mathcal{A}(F_t) \in \mathbb{R}^{D}.
\end{equation}

To that end, we could adaptively integrate action-shared semantics into conditioning, while avoiding the computational overhead of repeated diffusion processes. 
The aggregated condition is subsequently used to guide the denoising network to progressively generate foreground knowledge.

\begin{algorithm}[!t]
    \caption{BSD Training Procedure}
    \label{train}
    \renewcommand{\algorithmicrequire}{\textbf{\textcolor{red!60}{def}}}
     \renewcommand{\algorithmicensure}{\textbf{Output:}}
    \begin{algorithmic}[0]  
        \REQUIRE \textbf{train\_bsd} ($video\_feats$, $foreground\_feats$, $T$, $c$): 
        
        \vspace{0.4em}
        \STATE \hspace{1.8em}  \textcolor{blue!70}{ \# 1. Compute video / foreground representations}: \\
        \hspace{1.8em} $h_v = average\_pool(video\_feats)$; \hspace{3.0em} \textcolor{blue!70}{$\# [B,D]$ } \\
        \hspace{1.8em} $h_f = average\_pool(foreground\_feats)$; \hspace{0.2em} \textcolor{blue!70}{$\# [B,D]$ } \\
        
        \vspace{0.4em}
        \STATE \hspace{1.8em}  \textcolor{blue!70}{ \# 2. Compute Background residual}: \\
        \hspace{1.8em} $e = h_v - h_f$;
        
        \vspace{0.4em}
        \STATE \hspace{1.8em}  \textcolor{blue!70}{ \# 3. Sample diffusion timestep}: \\
        \hspace{1.8em} $t = randint(1, T)$;

        \vspace{0.4em}
        \STATE \hspace{1.8em}  \textcolor{blue!70}{ \# 4. Sample Gaussian noise}: \\
        \hspace{1.8em} $eps = normal(mean=0, std=1, shape=[B, D])$;

        \vspace{0.4em}
        \STATE \hspace{1.8em}  \textcolor{blue!70}{ \# 5. Forward residual diffusion}: \\
        \hspace{1.8em} $h_t = h_v - \gamma[t] * e + sqrt(\gamma[t]) * \sigma * eps$;  \hspace{0.1em} \textcolor{blue!70}{$\# [B,D]$ } \\

        \vspace{0.4em}
        \STATE \hspace{1.8em}  \textcolor{blue!70}{ \# 6. Foreground denoising prediction}: \\
        \hspace{1.8em} $h_f^{pred} = \phi_\theta(h_t, t, c)$; \hspace{6.8em} \textcolor{blue!70}{$\# [B,D]$ } \\

        \vspace{0.4em}
        \STATE \hspace{1.8em}  \textcolor{blue!70}{ \# 7. Denoising loss}: \\
        \hspace{1.8em} $loss = reconstrction\_loss(h_f^{pred}, h_f)$; \\

        \vspace{0.4em}
        \hspace{1.8em} \textbf{\textcolor{red!60}{Return}} $loss$;

    \end{algorithmic}
\end{algorithm}

\subsection{Background-Suppress Denoising}
To enable precise transformation from video representations to foreground knowledge, we aim to explicitly suppresses redundant background semantics through a diffusion denoising process conditioned on the action semantic condition $c$ generated above, as illustrated in Algorithm~\ref{train} and ~\ref{infer}.
Specifically, we first apply average pooling over all video frame features to obtain the video representation $h_{v}$. 
Likewise, the foreground frames are also averaged to derive the ground-truth foreground knowledge $h_f$. 
Then, the background residual can be defined as:
\begin{equation}
e=h_v-h_f,
\end{equation}
which captures redundant action-irrelevant information and need to be progressively removed via denoising process.

\vspace{0.12cm}

\noindent
\textbf{Forward Process.} 
During training, we construct a forward diffusion process that gradually injects background information into the target foreground knowledge, transforming $h_f$ to $h_{v}$. 
The residual injection strength is controlled by a monotonically increasing coefficient schedule $\{\gamma_t\}_{t=1}^{T}$, where $\gamma_1 \to 0$ and $\gamma_T \to 1$. 
The forward transition at step $t$ is defined as:
\begin{equation}
q(h_t|h_{t-1};h_v)=\mathcal{N}(h_t; h_{t-1} + \alpha_te, \alpha_t\sigma^2I),
\end{equation}
where $h_t$ denotes the intermediate semantic state ($h_0=h_f$, $h_T=h_v$), $\alpha_1=\gamma_1$, and $\alpha_t=\gamma_t-\gamma_{t-1}$ for $t>1$. 
$I$ is the identity matrix and $\sigma$ denotes the standard deviation.
By leveraging the reparameterization technique, the final forward process is simplified to:
\begin{equation}
q(h_t|h_v)=\mathcal{N}(h_t;h_v-\gamma_te, \gamma_t\sigma^2I).
\end{equation}

\noindent
\textbf{Reverse Process.} 
Conditioned on the action semantic condition $c$, the reverse process aims to iteratively remove the background residuals and recover the foreground knowledge. 
Unlike conventional diffusion models that initialize from pure Gaussian noise, our reverse process starts from the video representation $h_v$, leading to more accurate and stable semantic refinement. 
The conditional distribution for a single-step reverse transition is given by:
\begin{equation}
q(h_{t-1}|h_t;h_v) = \mathcal{N}(h_{t-1}; \frac{\gamma_{t-1}}{\gamma_t}h_t + \frac{\alpha_{t}}{\gamma_t}h_f, \frac{\gamma_{t-1}}{\gamma_t}\alpha_t\sigma^2I).
\end{equation}

To make the reverse process learnable, following prior diffusion-based methods~\cite{li2023momentdiff, liu2023conditional}, we parameterize the mean using a denoising network $\phi_{\theta}$.
The parameterized reverse mean is defined as:
\begin{equation}
\mu_{\theta}(h_t;t;c)=\frac{\gamma_{t-1}}{\gamma_t}h_t + \frac{\alpha_t}{\gamma_t}\phi_{\theta}(h_t;t;c).
\end{equation}

Accordingly, the reverse update step is:
\begin{equation}
h_{t-1} = \mu_{\theta}(h_t;t;c) + \lambda_t\epsilon,
\end{equation}
where $\lambda_t=\sqrt{\frac{\gamma_{t-1}}{\gamma_t}\alpha_t\sigma}$, and $\epsilon \sim \mathcal{N}(0,I)$ is the Gaussian noise.

\vspace{0.12cm}

\noindent
\textbf{Denoising Training.}
To train the denoising network $\phi_{\theta}$, we uniformly sample a timestep $t\in \{1,...,T\}$ and minimize the reconstruction error between the denoised output $h_t^{pred}$ and the ground-truth foreground feature $h_f$ by:
\begin{equation}
\mathcal{L}_{df}=||\phi_{\theta}(h_t,t,c)-h_f||^2.
\end{equation}
This diffusion denoising loss encourages the network to progressively align the intermediate representations at each timestep with the target foreground features, leading to more stable optimization and more accurate foreground knowledge generation.

\begin{algorithm}[!t]
    \caption{BSD Inference Procedure}
    \label{infer}
    \renewcommand{\algorithmicrequire}{\textbf{\textcolor{red!60}{def}}}
     \renewcommand{\algorithmicensure}{\textbf{Output:}}
    \begin{algorithmic}[0]  
        \REQUIRE \textbf{infer\_bsd} ($video\_feats$, $T$, $c$): 
        
        \vspace{0.4em}
        \STATE \hspace{1.8em}  \textcolor{blue!70}{ \# 1. Initialize from video representations}: \\
        \hspace{1.8em} $h_v = average\_pool(video\_feats)$; \hspace{2.5em} \textcolor{blue!70}{$\# [B,D]$ } \\
        \hspace{1.8em} $h_t = h_v$; \hspace{12.8em} \textcolor{blue!70}{$\# [B,D]$ } \\
        
        \vspace{0.4em}
        \STATE \hspace{1.8em}  \textcolor{blue!70}{ \# 2. Reverse diffusion denoising}: \\
        \hspace{1.8em} \textcolor{red!60}{for} \hspace{0.1em} $t$ \hspace{0.1em} in \hspace{0.1em} \textcolor{red!60}{reversed} ( range ($1$, $T + 1$)): \\
        \hspace{3.0em} \textcolor{blue!70}{ \# Predict foreground knowledge at timestep $t$}: \\
        \hspace{3.0em} $h_t^{pred} = \phi_\theta(h_t, t, c^{train})$; \hspace{5.3em} \textcolor{blue!70}{$\# [B,D]$ } \\
        \hspace{3.0em} \textcolor{blue!70}{ \# Compute reverse mean}: \\
        \hspace{3.0em} 
        $\mu = (\gamma[t - 1] / \gamma[t]) * h_t \
             + (\alpha[t] / \gamma[t]) * h_t^{pred}$; \\
        \hspace{3.0em} \textcolor{blue!70}{ \# Sample noise}: \\
        \hspace{3.0em} \textcolor{red!60}{if} \hspace{0.1em}  $t > 1$: \\
        \hspace{4.0em} $eps = normal(mean=0, std=1, shape=[B, D])$; \\
        \hspace{3.0em} \textcolor{red!60}{else}: \\
        \hspace{4.0em} $eps = 0$;
        \STATE \hspace{3.0em}  \textcolor{blue!70}{ \# Reverse update}: \\
        \hspace{3.0em} $h_t = \mu + sqrt((\gamma[t - 1] / \gamma[t]) * \alpha[t]) * \sigma * eps$; \\

        \vspace{0.4em}
        \STATE \hspace{1.8em}  \textcolor{blue!70}{ \# 3. Final predicted foreground knowledge}: \\
        \hspace{1.8em} $h_f^{pred} = h_t$

        \vspace{0.4em}
        \hspace{1.8em} \textbf{\textcolor{red!60}{Return}} $h_f^{pred}$;

    \end{algorithmic}
\end{algorithm}

\subsection{Foreground-Prompt Alignment}
Prior OV-TAD methods often overlook the semantic gap between visual and textual modalities and are susceptible to interference from redundant background content.
To address these limitations, we leverage extracted foreground knowledge as explicit guidance for action–video alignment. 
Specifically, we incorporate the foreground knowledge as additional prompt tokens into action labels, then align with the corresponding video segments for action classification and localization. 

\vspace{0.12cm}

\noindent
\textbf{Multi-modal Interaction.}
Typically, video and text features are interacted directly to enhance the model’s focus on action-relevant segments. 
However, the significant semantic differences between the two modalities often make this interaction challenging. 
To alleviate this issue, we first embed the extracted foreground knowledge $h_f^{pred}$ as prompt tokens into the action labels:
\begin{equation}
F_t^{'}=Proj([F_{t}^{specific};h_f^{pred}]) \subseteq \mathbb{R}^{C \times D },
\end{equation}
where $[\cdot;\cdot]$ denotes concatenation along the token dimension, and $\mathrm{Proj}(\cdot)$ is a projection operator that maps $F_t^{\prime}$ to dimension $D$.

We then interact the foreground-integrated text representations $F_t^{'}$ with video features $F_v$ through the multi-head cross-attention mechanism to extract action-relevant segments:
\begin{equation}
F_v^{'} = Softmax(\frac{Q(F_v)K(F_{t}^{'})^{\top}}{\sqrt{D}})V(F_{t}^{'}),
\end{equation}
where $Q$, $K$ and $V$ are learnable linear projections.
By injecting foreground knowledge into text representations, our method enables more effective extraction of discriminative action-relevant cues from videos, consequently improving cross-modal alignment.

\vspace{0.12cm}

\noindent
\textbf{Action Classification.}
In the OV-TAD setting, actions are classified by computing the similarity between action label embeddings and video representations. 
However, since action labels are often concise and insufficient to capture the rich contextual semantics of videos, direct alignment can lead to misclassification.
In contrast, we align updated video representations $F_v^{'}$ with foreground knowledge-integrated text representations $F_t^{'}$ as:
\begin{equation}
C_{cls} = F_{v}^{'} F_{t}^{' \top} \in \mathbb{R}^{N_v \times C},
\end{equation}
where $C_{cls}$ denotes the probability over all categories at each segment.
Finally, we employ the cross-entropy loss to supervise the action classification by:
\begin{equation}
 \mathcal{L}_{cls} = CrossEntropy(C_{cls}, G_{cls}),
\end{equation}
where $G_{cls} \in \mathbb{R}^{N_v \times C}$ is the one-hot ground-truth label.
By enriching action label semantics with foreground knowledge, we effectively bridges the semantic gap between videos and actions, resulting in more precise cross-modal alignment for action classification.

\vspace{0.12cm}

\noindent
\textbf{Action Localization.} 
We predict the distances $d_s^{start}$ and $d_s^{end}$ from each temporal segment $s$ to the action start and end boundaries, by feeding the action-enhanced video representations $F_v^{'}$ into a foreground-aware head and a regression head, respectively. 
These predictions are supervised using a foreground-aware loss $\mathcal{L}_{fg}$ and a DIoU-based localization loss $\mathcal{L}_{loc}$, following~\cite{lee2024text}. 

\subsection{Training and Inference}
\textbf{Training Loss.}
Combining the diffusion denoising loss $\mathcal{L}_{df}$, the action classification loss $\mathcal{L}_{cls}$, the foreground-aware loss $\mathcal{L}_{fg}$ and the action localization loss $\mathcal{L}_{loc}$ together, the overall objective is formulated as:
\begin{equation}
\mathcal{L} = \mathcal{L}_{df} + \mathcal{L}_{cls} + \mathcal{L}_{fg}  + \mathcal{L}_{loc}.
\end{equation}

\noindent
\textbf{Inference.}
At test time, we first feed action labels from the test split into the Semantic-Unify Conditioning module to obtain the action semantic condition $c$. 
The test videos are then processed by the Background-Suppress Denoising module, which progressively removes redundant background semantics through a $T$-step diffusion process conditioned on $c$, yielding the foreground knowledge $h_{f}^{pred}$. 
The extracted foreground knowledge is subsequently embedded as prompt tokens into the action labels and interacts with video representations via the Foreground-Prompt Alignment module to predict $(d_s^{start}, d_s^{end}, p(c_s))$ at each temporal segment $s$, where $d_s^{start}$ and $d_s^{end}$ denote the distances from $s$ to the start and end boundaries of an action instance, respectively.
Finally, redundant proposals are suppressed using Soft-NMS~\cite{bodla2017soft} to produce the final action predictions.

\section{Experiments}

\subsection{Experimental Settings}
\textbf{Datasets.}
We conduct experiments on two widely used open-vocabulary temporal action detection (OV-TAD) benchmarks: THUMOS14~\cite{idrees2017thumos} and ActivityNet v1.3~\cite{caba2015activitynet}. \textbf{THUMOS14} contains 200 training videos and 213 testing videos spanning 20 sports actions. \textbf{ActivityNet v1.3} includes 19,994 videos covering 200 daily action categories.
Following prior work~\cite{nag2022zero, lee2024text, phan2024zeetad}, we evaluate under two open-vocabulary splits: a 50\%–50\% setting (training on 50\% of the action categories and testing on the remaining 50\%) and a 75\%–25\% setting (training on 75\% of the categories and testing on the remaining 25\%). To ensure statistical reliability, we adopt 10 random category splits and report the averaged performance.

\vspace{0.12cm}

\noindent
\textbf{Evaluation Metric.} 
We adopt mean Average Precision (mAP) as the evaluation metric, computed by averaging the precisions at different temporal intersection over union (tIoU) thresholds. For THUMOS14, the tIoU thresholds range from 0.3 to 0.7 with a step size of 0.1 ([0.3:0.1:0.7]). For ActivityNet v1.3, the tIoU thresholds range from 0.5 to 0.95 at intervals of 0.05 ([0.5:0.05:0.95]).

\vspace{0.12cm}

\noindent
\textbf{Baselines.} 
We compare our approach against ten state-of-the-art TAD baselines, including three methods originally developed for closed-set TAD but adapted to the OV-TAD setting: TriDet~\cite{shi2023tridet}, DyFADet~\cite{yang2024dyfadet}, and DiGIT~\cite{kim2025digit}, as well as seven methods specifically designed for OV-TAD: EffPrompt~\cite{ju2022prompting}, STALE~\cite{nag2022zero}, DeTAL~\cite{li2024detal}, CSP~\cite{wangconcept}, ZEETAD~\cite{phan2024zeetad}, STOV~\cite{hyun2025exploring}, and Ti-FAD~\cite{lee2024text}. Closed-set methods are adapted to the open-vocabulary setting following the protocol in~\cite{li2024detal, wangconcept}, which partitions the action label space into disjoint sets for training and testing.

\vspace{0.12cm}

\noindent
\textbf{Implementation Details.} 
Following prior works~\cite{nag2022zero, lee2024text, hyun2025exploring}, we construct visual features by concatenating the RGB and optical flow representations extracted from the two-stream I3D network. For textual features, we employ the frozen pre-trained CLIP ViT-B/16 model~\cite{radford2021learning} as the text encoder.
Our model is trained for 19 epochs on THUMOS14 with a batch size of 2, and for 10 epochs on ActivityNet v1.3 with a batch size of 16. we adopt the Adam optimizer with a linear warm-up for the first 5 epochs. 
The initial learning rate is set to 0.0001. 
For the architecture of denoising network, we employ 3 CrossDiT blocks, each with 2 attention heads, a hidden dimension of 512, and 3 transformer layers. The denoising step is set to 8. 

\begin{table*}[t]
\centering
\caption{Performance comparison with the state-of-the-art methods on THUMOS14 and ActivityNet v1.3. * indicates closed-set TAD methods adapted to the OV-TAD setting. The best and second-best results are highlighted in \textbf{\textcolor{red}{Red}} and \textbf{\textcolor{blue}{Blue}}, respectively.}
\resizebox{\textwidth}{!}{
\begin{tabular}{c c c c c c c c c c c c} 
\hline
\multirow{2}{*}{Data Split} & \multirow{2}{*}{Methods} & \multicolumn{6}{c}{THUMOS14} & \multicolumn{4}{c}{ActivityNet v1.3} \\
\cmidrule(lr){3-8} \cmidrule(lr){9-12}
 &  & 0.3 & 0.4 & 0.5 & 0.6 & 0.7 & Avg mAP(\%) & 0.5 & 0.75 & 0.95 & Avg mAP(\%) \\
\hline
\multirow{11}{*}{\shortstack{50\% Seen \\ 50\% Unseen}} 
& TriDet* (ICCV'23) & 15.2 & 13.2 & 10.8 & 7.9 & 5.2 & 10.5 & 19.1 & 11.5 & 1.1 & 11.4 \\
& DyFADet* (ECCV'24) & 17.5 & 14.9 & 12.2 & 9.2 & 5.7 & 11.9 & 23.8 & 14.2 & 1.8 & 13.6 \\
& DiGIT* (CVPR'25) & 19.1 & 16.2 & 13.5 & 10.3 & 6.1 & 13.0 & 27.5 & 17.3 & 2.3 & 16.0 \\
& EffPrompt (ECCV'22) & 37.2 & 29.6 & 21.6 & 14.0 & 7.2 & 21.9 & 32.0 & 19.3 & 2.9 & 19.6 \\
& STALE (ECCV'22) & 38.3 & 30.7 & 21.2 & 13.8 & 7.0 & 22.2 & 32.1 & 20.7 & 5.9 & 20.5 \\
& DeTAL (TPAMI'24) & 38.3 & 32.3 & 24.4 & 16.3 & 9.0 & 24.1 & 34.4 & 23.0 & 4.0 & 22.4 \\
& CSP (JCST'25) & 41.2 & 33.4 & 24.8 & 17.3 & 10.9 & 25.5 & 38.4 & 26.4 & 5.2 & 25.7 \\
& ZEETAD (WACV'24) & 45.2 & 38.8 & 30.8 & 22.5 & 13.7 & 30.2 & 39.2 & 25.7 & 3.1 & 24.9 \\
& STOV (WACV'25) & 56.3 & - & 34.4 & - & 11.3 & 34.0 & 48.4 & 28.7 & - & 27.9 \\
& Ti-FAD (NeurlPS'24) & \textcolor{blue}{\underline{57.0}} & \textcolor{blue}{\underline{51.4}} & \textcolor{blue}{\underline{43.3}} & \textcolor{blue}{\underline{33.0}} & \textcolor{blue}{\underline{21.2}} & \textcolor{blue}{\underline{41.2}} & \textcolor{blue}{\underline{50.6}} & \textcolor{blue}{\underline{32.2}} & \textcolor{blue}{\underline{5.2}} & \textcolor{blue}{\underline{32.0}} \\
\cmidrule(lr){2-12}
& \textbf{Ours} & \textcolor{red}{\textbf{69.6}} & \textcolor{red}{\textbf{63.0}} & \textcolor{red}{\textbf{54.1}} & \textcolor{red}{\textbf{42.8}} & \textcolor{red}{\textbf{29.4}} & \textcolor{red}{\textbf{51.8}}& \textcolor{red}{\textbf{53.3}} & \textcolor{red}{\textbf{35.5}} & \textcolor{red}{\textbf{7.9}} & \textcolor{red}{\textbf{35.7}} \\
\hline
\multirow{11}{*}{\shortstack{75\% Seen \\ 25\% Unseen}} 
& TriDet* (ICCV'23) & 25.9 & 22.5 & 18.2 & 13.1 & 6.2 & 17.2 & 25.5 & 15.2 & 2.0 & 15.3 \\
& DyFADet* (ECCV'24) & 27.6 & 23.9 & 19.4 & 13.8 & 6.7 & 18.3 & 28.9 & 17.6 & 2.5 & 17.1 \\
& DiGIT* (CVPR'25) & 29.0 & 25.2 & 20.5 & 14.3 & 7.0 & 19.2 & 32.2 & 19.7 & 2.9 & 18.8 \\
& EffPrompt (ECCV'22) & 39.7 & 31.6 & 23.0 & 14.9 & 7.5 & 23.3 & 37.6 & 22.9 & 3.8 & 23.1 \\
& STALE (ECCV'22) & 40.5 & 32.3 & 23.5 & 15.3 & 7.6 & 23.8 & 38.2 & 25.2 & 6.0 & 24.9 \\
& DeTAL (TPAMI'24) & 39.8 & 33.6 & 25.9 & 17.4 & 9.9 & 25.3 & 39.3 & 26.4 & 5.0 & 25.8 \\
& CSP (JCST'25) & 42.7 & 35.5 & 26.4 & 18.5 & 12.0 & 27.0 & 41.1 & 28.8 & 7.4 & 28.1 \\
& ZEETAD (WACV'24) & 61.4 & 53.9 & 44.7 & 34.5 & 20.5 & 43.2 & 51.0 & 33.4 & 5.9 & 32.5 \\
& STOV (WACV'25) & 59.5 & - & 37.5 & - & 12.5 & 36.9 & 52.0 & 30.6 & - & 30.1 \\
& Ti-FAD (NeurlPS'24) & \textcolor{blue}{\underline{64.0}} & \textcolor{blue}{\underline{58.5}} & \textcolor{blue}{\underline{49.7}} & \textcolor{blue}{\underline{37.7}} & \textcolor{blue}{\underline{24.1}} & \textcolor{blue}{\underline{46.8}} & \textcolor{blue}{\underline{53.8}} & \textcolor{blue}{\underline{34.8}} & \textcolor{blue}{\underline{7.0}} & \textcolor{blue}{\underline{34.7}} \\
\cmidrule(lr){2-12}
& \textbf{Ours} & \textcolor{red}{\textbf{73.9}} & \textcolor{red}{\textbf{67.1}} & \textcolor{red}{\textbf{57.8}} & \textcolor{red}{\textbf{45.1}} & \textcolor{red}{\textbf{30.6}} & \textcolor{red}{\textbf{54.9}} & \textcolor{red}{\textbf{56.3}} & \textcolor{red}{\textbf{37.3}} & \textcolor{red}{\textbf{9.5}} & \textcolor{red}{\textbf{37.1}}\\
\hline
\end{tabular}
}
\label{performance comparison}
\end{table*}

\subsection{Comparison with State-of-the-Arts}
We evaluate the performance of our proposed method by comparing it with state-of-the-art OV-TAD approaches as well as adapted fully-supervised TAD baselines. 
As in Table~\ref{performance comparison}, our method consistently outperforms existing methods on both the THUMOS14 and ActivityNet1.3 datasets, achieving higher mAP scores across all evaluated tIoUs. 
For example, under the 50\% Seen / 50\% Unseen split, our method achieves a 25.7\% and 11.6\% relative lift in average mAP over SOTA competitor Ti-FAD on two benchmarks, respectively. 
This is consistent with the 75\% Seen / 25\% Unseen split.
These gains stem from the use of foreground knowledge, which is generated by progressively removing background information from video features via a diffusion-based denoiser and injected as prompt tokens into text representations. 
This design effectively narrows the semantic gap between action labels and foreground video segments, resulting in improved detection performance on unseen categories.

\begin{figure*}[t!]
\centering
  \includegraphics[width=\textwidth]{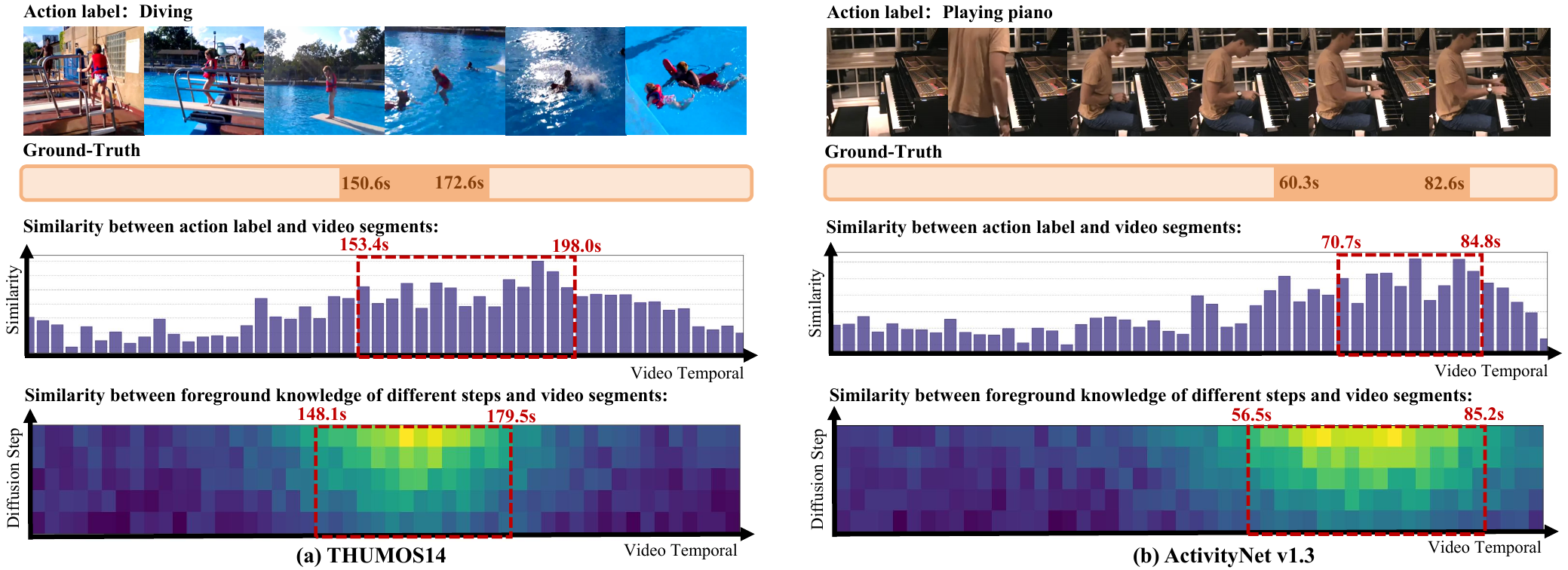}
  \caption{Similarity between action label and video segments,
as well as between foreground knowledge of different steps and video segments. }
  \label{heatmap}
\end{figure*}

\subsection{Ablation Studies}
\noindent
\textbf{Analysis of extracted foreground knowledge.} 
To evaluate the quality of the foreground knowledge extracted by the Background-Suppress Denoising (BSD) module, we randomly select one video from THUMOS14 and ActivityNet v1.3, respectively. 
Besides, we also visualize the similarity between video segments and corresponding action labels, as well as between foreground knowledge at different diffusion steps and video segments. 
As shown in Figure~\ref{heatmap}, the final foreground knowledge achieves more precise alignment than direct action label–video matching. 
The reason is that the extracted foreground knowledge is gradually refined via iterative diffusion denoising, which incrementally removes background residuals and concentrates semantic attention on action-relevant temporal regions. 
This progressive refinement enables the model to disentangle foreground semantics from complex video context, leading to more accurate alignment with action-relevant segments.

\vspace{0.12cm}

\noindent
\textbf{Analysis of each component.}
Table~\ref{component analysis} presents the ablations on three core modules: SUC, BSD and FPA. 
The baseline model (Row 1) directly aligns action labels with video features for detection.
Rows 2, 3 and 4 progressively incorporate FPA, BSD and SUC to assess their individual contributions.
Specifically, Row 2 introduces the FPA module, which injects the averaged video representation as prompt tokens into text-representations to facilitate action–video interaction.
Row 3 further incorporates the BSD module, where action label embeddings are averaged to form action semantics that guide the diffusion denoising process.
Clearly, incorporating BSD (Row 3) shows substantial additional gain, with averaged improvements of 12.6\% and 7.3\% over Rows 1 and 2, respectively, highlighting the effectiveness of diffusion-generated foreground knowledge in reducing the semantic gap between action labels and video segments. 
Furthermore, adding SUC (Row 4) consistently boosts performance by an additional 3.9\%, demonstrating that unifying shared and action-specific semantics yields more informative conditions for guiding the diffusion denoising process.

\begin{table}[t!]
  \normalsize
  \caption{Analysis of each components on THUMOS14.}
  \begin{tabularx}{\columnwidth}{ >{\raggedright\arraybackslash}p{0.13\linewidth}  >{\centering\arraybackslash}X  >
  {\centering\arraybackslash}X  >
  {\centering\arraybackslash}X  >{\centering\arraybackslash}p{0.2\linewidth}  >{\centering\arraybackslash}p{0.2\linewidth}}
  \hline
   \multirow{2}{*}{Method} & \multirow{2}{*}{FPA} & \multirow{2}{*}{BSD} & \multirow{2}{*}{SUC} & \multicolumn{2}{c}{mAP@AVG} \\
    \cmidrule(lr){5-6}
   & & & & 50\%-50\% & 75\%-25\% \\
   \hline
   Baseline & \ding{55} & \ding{55} & \ding{55} & 42.5 & 48.9 \\
   \hline
   \multirow{3}{*}{Ours} & \ding{51} & \ding{55} & \ding{55} & 45.4 & 50.5\\
   & \ding{51} & \ding{51} & \ding{55} & 49.6 & 53.1\\
   & \ding{51} & \ding{51} & \ding{51} & \textcolor{red}{\textbf{51.8}} & \textcolor{red}{\textbf{54.9}}\\
  \hline
  \end{tabularx}
  \label{component analysis}
\end{table}

\begin{table}[!]
  \normalsize
   \caption{Analysis of different conditioning methods in SUC module under the 50\% seen / 50\% unseen split on THUMOS14.}
  \begin{tabularx}{\columnwidth}{ >{\raggedright\arraybackslash}p{0.33\linewidth}  >{\centering\arraybackslash}X  >{\centering\arraybackslash}X  >{\centering\arraybackslash}X  >{\centering\arraybackslash}X >
  {\centering\arraybackslash}X}
  
  \hline
   \multirow{2}{*}{Condition Type} & \multicolumn{4}{c}{mAP@tIOU (\%)} & \multirow{2}{*}{Time (s)} \\
   \cmidrule(lr){2-5}  
   & 0.3 & 0.5 & 0.7 & Avg & \\
   \hline
   w/o condition & 65.6 & 51.6 & 24.7 & 47.8 & 37.2\\
   \hline
    Per-action & 67.9 & 52.1 & 25.0 & 48.9 & 51.1\\
    Shared Only & 66.7 & 51.6 & 24.6 & 48.2 & 38.7\\
    Action-Specific Only & 67.9 & 52.8 & 26.9 & 49.6 & 37.9\\
    \hline
    \textbf{SUC (Ours)} & \textcolor{red}{\textbf{69.6}} & \textcolor{red}{\textbf{54.1}} & \textcolor{red}{\textbf{29.4}} & \textcolor{red}{\textbf{51.8}} & 40.4 \\
    \hline
  \end{tabularx}
  \label{BSCG condition}
\end{table}

\vspace{0.12cm}

\noindent
\textbf{Analysis of SUC Module.} 
To evaluate the effectiveness of the proposed SUC module, we first compare different conditioning methods. 
As shown in Table~\ref{BSCG condition}, we introduce four variants:  1) \emph{w/o Condition}, which removes action semantic conditioning from the diffusion denoising process; 2) \emph{Per-action}, which independently conditions the denoising process on each action category; 3) \emph{Shared Only}, which performs conditioning using action-shared semantics summarized via LLM prompting; and 4) \emph{Action-Specific Only}, which performs conditioning using action-specific semantics aggregated by the Foreground Semantic Aggregator. 
We can draw four key observations: 1) Conditioned variants consistently outperform the variant without semantic conditioning, confirming that action conditions are crucial for guiding the diffusion process toward discriminative foreground knowledge. 
2) Per-action Diffusion incurs substantially higher inference time with only limited performance gains, highlighting its inefficiency in practice. 
3) Using either shared semantics or action-specific semantics alone results in inferior performance, indicating that neither component by itself provides sufficient guidance for effective foreground semantic conditioning. 
4) Our SUC achieves the best overall performance with limited inference overhead, demonstrating that unifying action-shared and action-specific semantics enables more informative and efficient semantic conditioning.

Furthermore, we examine the impact of different strategies for unifying action-shared and action-specific semantics in the SUC module.
Specifically, we compare with two variants: \emph{Addition}, which directly sums the action-shared and action-specific text embeddings, and \emph{Concatenation}, which concatenates the two embeddings along the feature dimension followed by a linear projection to the original dimensionality.
As reported in Table~\ref{BSRG aggregation}, both variants yield inferior performance compared to our proposed Fogreground Semantic Aggregator (FSA).
This demonstrates that FSA achieves more effective semantic aggregation by modeling contextual interactions between action-shared and action-specific semantics and adaptively integrating them through a Transformer.

\begin{table}[t!]
  \normalsize
   \caption{Analysis of different unify strategies in SUC module under the 50\% seen / 50\% unseen split on THUMOS14. }
  \begin{tabularx}{\columnwidth}{ >{\raggedright\arraybackslash}p{0.28\linewidth}  >{\centering\arraybackslash}X  >{\centering\arraybackslash}X  >{\centering\arraybackslash}X  >{\centering\arraybackslash}X}
  \hline
   \multirow{2}{*}{Unify Strategy} & \multicolumn{4}{c}{mAP@tIOU (\%)} \\
   \cmidrule(lr){2-5}
   & 0.3 & 0.5 & 0.7 & Avg \\
  \hline
   Addition & 68.6 & 53.2 & 28.7  & 51.0 \\
   concatenation & 68.2 & 52.9 & 28.3 & 50.5 \\
   \hline
   \textbf{FSA(ours)} & \textcolor{red}{\textbf{69.6}} & \textcolor{red}{\textbf{54.1}} & \textcolor{red}{\textbf{29.4}} &  \textcolor{red}{\textbf{51.8}}\\
    \hline
  \end{tabularx}
  \label{BSRG aggregation}
\end{table}

\vspace{0.12cm}

\noindent
\textbf{Analysis of Different LLM Backbone.} 
To evaluate the robustness of SUC module across LLM backbones, we compare four widely adopted LLMs: Qwen3~\cite{yang2025qwen3}, DeepSeek-v3~\cite{liu2024deepseek}, GPT-4~\cite{achiam2023gpt}, and GPT-4o~\cite{hurst2024gpt}. 
The results are presented in Table~\ref{LLM selection}. We observe that overall performance remains relatively stable across different LLM backbones, suggesting that the proposed SUC module is robust to backbone choice, which is desirable for practical deployment.

\begin{table}[t!]
  \normalsize
   \caption{Analysis of Different LLM Backbone in SUC module under the 50\% seen / 50\% unseen split on THUMOS14.}
  \begin{tabularx}{\columnwidth}{ >{\raggedright\arraybackslash}p{0.28\linewidth}  >{\centering\arraybackslash}X  >{\centering\arraybackslash}X  >{\centering\arraybackslash}X  >{\centering\arraybackslash}X}
  \hline
   \multirow{2}{*}{LLM Backbone} & \multicolumn{4}{c}{mAP@tIOU (\%)} \\
   \cmidrule(lr){2-5}
   & 0.3 & 0.5 & 0.7 & Avg \\
  \hline
   Qwen3 & 68.6 & 53.4 & 28.9 &  51.0 \\
   Deepseek v3 & 69.1 & 53.7 & 29.0 & 51.3 \\
   GPT-4 & 68.9 & 53.6 & 29.2 & 51.3 \\
   \textbf{GPT-4o} & \textcolor{red}{\textbf{69.6}} & \textcolor{red}{\textbf{54.1}} & \textcolor{red}{\textbf{29.4}} &  \textcolor{red}{\textbf{51.8}}\\
    \hline
  \end{tabularx}
  \label{LLM selection}
\end{table}

\noindent
\textbf{Analysis of BSD Module.} 
To evaluate the effectiveness of the proposed BSD module, we first compare different diffusion models. The results are shown in Table~\ref{different diffusion models}. 
DDPM~\cite{ho2020denoising} and DDIM~\cite{song2020denoising} are two representative diffusion sampling strategies: \emph{DDPM} employs a stochastic reverse process with dense sampling, whereas \emph{DDIM} adopts a deterministic formulation that enables efficient inference with fewer steps.
Despite their differences in sampling strategy, both methods initialize the reverse process from pure Gaussian noise, making it challenging to generate foreground knowledge with rich contextual semantics.
In contrast, our BSD module initializes the diffusion process from video representation, substantially reducing generation difficulty and leading to improved performance.

We also analyze the impact of different total timesteps in the diffusion process. 
As shown in Table~\ref{total steps}, w/o diffusion removes the diffusion-based denoising process and the resulting foreground knowledge. 
Two observations can be drawn: 
1) All diffusion-based variants consistently outperform w/o diffusion, indicating that diffusion-denoised foreground knowledge is crucial for effective action–video alignment. 
2) Increasing the number of diffusion steps initially improves detection performance, as iterative denoising enables progressively finer semantic refinement; However, when the number of steps exceeds 8, the performance begins to degrade. This is likely due to over-denoising, where excessive iterations may suppress discriminative details. In addition, more diffusion steps incur higher computational cost. 
Therefore, we adopt 8 diffusion steps in our implementation.

\begin{table}[t!]
  \normalsize
   \caption{Analysis of different diffusion models in BSD module under the 50\% seen / 50\% unseen split on THUMOS14. }
  \begin{tabularx}{\columnwidth}{ >{\raggedright\arraybackslash}p{0.28\linewidth}  >{\centering\arraybackslash}X  >{\centering\arraybackslash}X  >{\centering\arraybackslash}X  >{\centering\arraybackslash}X}
  \hline
   \multirow{2}{*}{Diffusion Model} & \multicolumn{4}{c}{mAP@tIOU (\%)} \\
   \cmidrule(lr){2-5}
   & 0.3 & 0.5 & 0.7 & Avg \\
  \hline
   DDPM (Step = 32) & 65.8 & 51.5 & 25.3 & 48.6 \\
   DDIM (Step = 8) & 66.0 & 52.2 & 26.8 & 49.3 \\
   \hline
   \textbf{BSD (Step = 8)} & \textcolor{red}{\textbf{69.6}} & \textcolor{red}{\textbf{54.1}} & \textcolor{red}{\textbf{29.4}} &  \textcolor{red}{\textbf{51.8}}\\
    \hline
  \end{tabularx}
  \label{different diffusion models}
\end{table}

\begin{table}[t!]
  \normalsize
   \caption{Analysis of total diffusion steps in BSD module under the 50\% seen / 50\% unseen split on THUMOS14.}
  \begin{tabularx}{\columnwidth}{ >{\raggedright\arraybackslash}p{0.25\linewidth}  >{\centering\arraybackslash}X  >{\centering\arraybackslash}X  >{\centering\arraybackslash}X  >{\centering\arraybackslash}X >
  {\centering\arraybackslash}X}
  
  \hline
   \multirow{2}{*}{Total Steps} & \multicolumn{4}{c}{mAP@tIOU (\%)} & \multirow{2}{*}{Time (s)} \\
   \cmidrule(lr){2-5}  
   & 0.3 & 0.5 & 0.7 & Avg & \\
   \hline
    w/o diffusion & 59.9 & 45.5 & 22.3 & 42.5 & 32.6 \\
    \hline
    Step = 2 & 65.2 & 51.3 & 27.1 & 48.6 & 34.8 \\
    Step = 4 & 66.8  & 51.5 & 27.5& 49.6 & 37.5\\
    \textbf{Step = 8} & \textcolor{red}{\textbf{69.6}} & \textcolor{red}{\textbf{54.1}} & \textcolor{red}{\textbf{29.4}} & \textcolor{red}{\textbf{51.8}} & 40.4 \\
    Step = 16 & 68.4 & 53.3 & 27.9 & 51.2 & 48.2 \\
    Step = 32 & 67.8 & 52.0 & 25.8 & 49.4 & 60.0 \\
    \hline
  \end{tabularx}
  \label{total steps}
\end{table}

\begin{figure*}[t]
\centering
  \includegraphics[width=\textwidth]{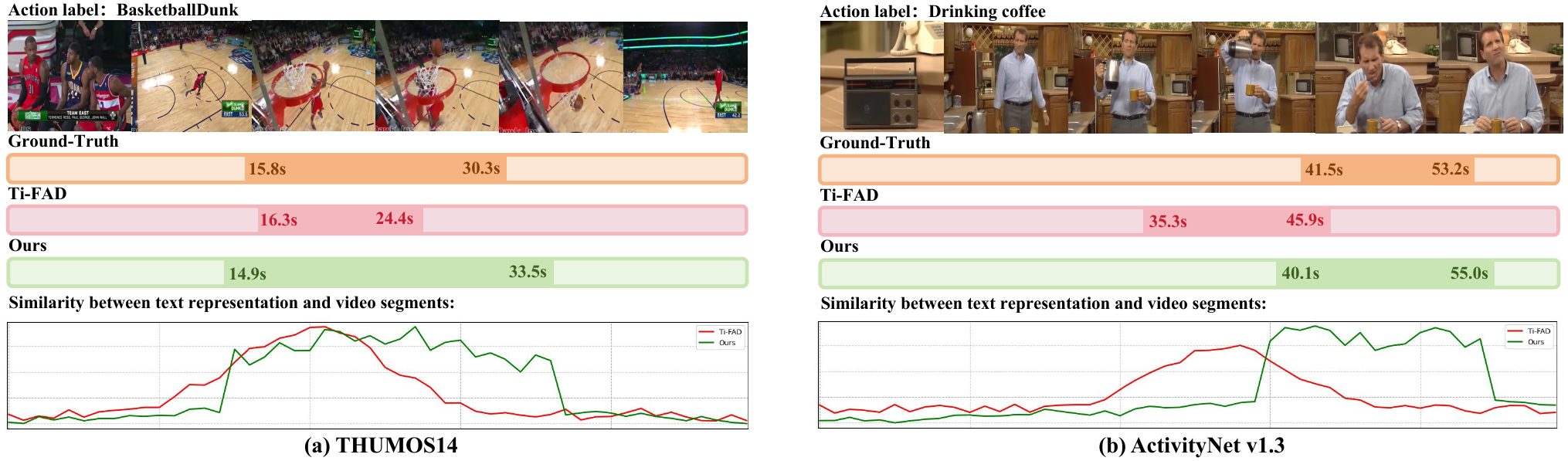}
  \caption{Visualization of the detection results and corresponding similarity between text representation and video segments. }
  \label{visualization}
\end{figure*}

\vspace{0.12cm}

\noindent
\textbf{Analysis of FPA Module.}
To evaluate the effectiveness of the FPA module, we replace the extracted foreground knowledge with three alternative prompt designs: 1) \emph{w/o Tokens}, which performs action–video alignment without any prompt tokens; 2) \emph{Learnable Tokens}, which injects randomly initialized, learnable embeddings as prompts; and 3) \emph{Video Tokens}, which directly use video features as prompt tokens for alignment.
The results are shown in Table~\ref{different ptompt token}. 
Two observations can be made: 
1) All prompt-based variants consistently outperform the w/o Tokens setting, indicating that introducing prompt tokens facilitates more effective action–video interaction by alleviating the cross-modal semantic gap. 
2) Compared with Learnable Tokens and Video Tokens, FPA achieves the best performance. 
Unlike learnable tokens that lack semantic information and video tokens that are dominated by background content, the extracted foreground knowledge provides explicit, action-relevant semantic cues, enabling more effective cross-modal alignment and leading to more accurate detection of unseen actions.

\begin{table}[!]
  \normalsize
   \caption{Analysis of Different Prompt Tokens in FPA module under the 50\% seen / 50\% unseen split on THUMOS14.}
  \begin{tabularx}{\columnwidth}{ >{\raggedright\arraybackslash}p{0.27\linewidth}  >{\centering\arraybackslash}X  >{\centering\arraybackslash}X  >{\centering\arraybackslash}X  >{\centering\arraybackslash}X}
  \hline
   \multirow{2}{*}{Prompt Token} & \multicolumn{4}{c}{mAP@tIOU (\%)} \\
   \cmidrule(lr){2-5}
   & 0.3 & 0.5 & 0.7 & Avg \\
  \hline
  w/o Tokens & 59.9 & 45.5 & 22.3 & 42.5 \\
  \hline
  Learnable Tokens & 62.1 & 46.5 & 23.7 & 44.0 \\
  Video Tokens & 63.8 & 47.0 & 24.8 & 45.4\\
   \hline
   \textbf{FPA (Ours)} & \textcolor{red}{\textbf{69.6}} & \textcolor{red}{\textbf{54.1}} & \textcolor{red}{\textbf{29.4}} &  \textcolor{red}{\textbf{51.8}} \\
    \hline
  \end{tabularx}
  \label{different ptompt token}
\end{table}

\vspace{0.12cm}

\subsection{Qualitative Results}
\begin{figure}[!]
\centering
  \includegraphics[width=1.0\linewidth]{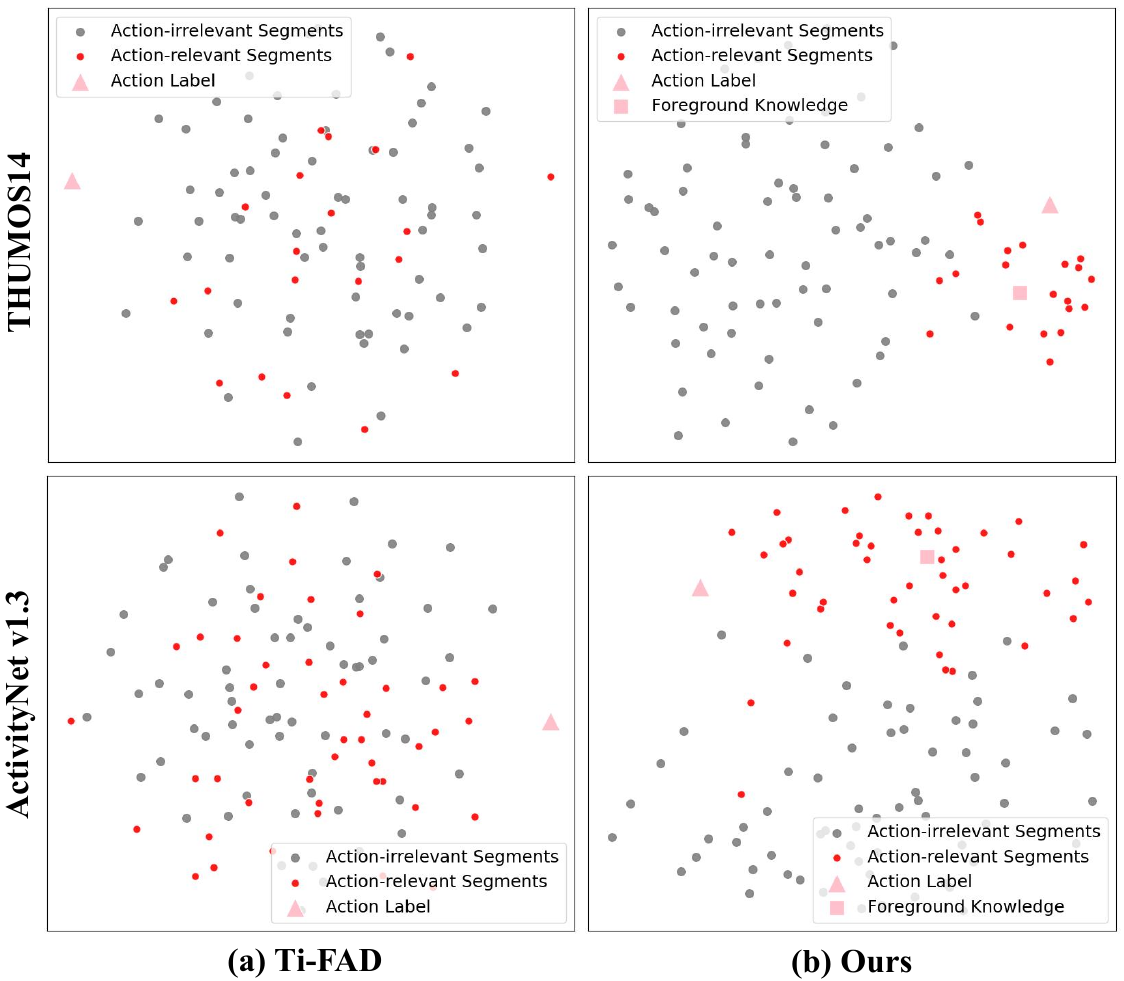}
  \caption{t-SNE visualization of the video segment embeddings and action label embeddings for Ti-FAD (a) and our proposed DFAlign (b) on THUMOS 14 and ActivityNet v1.3. }
  \label{TSNE}
\end{figure}

\textbf{Visualizations of the t-SNE comparison.}
Figure~\ref{TSNE} presents t-SNE visualization comparisons between Ti-FAD and our proposed DFAlign. 
Specifically, we randomly sample one video from the THUMOS14 and ActivityNet v1.3 test sets respectively, and project the corresponding segment-level video features and action label features into a 2D space using t-SNE~\cite{maaten2008visualizing}. We further visualize both the video representations and the extracted foreground knowledge produced by the proposed DFAlign.
We further visualize the video representations and the extracted foreground knowledge for our proposed DFAlign.
By analyzing the t-SNE results, DFAlign yields more compact and discriminative representations for action-relevant segments than Ti-FAD. 
These segments cluster more tightly around the learned foreground knowledge and lie closer to their corresponding action label embeddings. 
This indicates that the foreground knowledge extracted by the BSD module could serve as an effective intermediate semantic anchor between video and text representations, reducing the semantic gap and enhancing the discriminability of action-related segments, thereby enabling more precise cross-modal alignment.

\vspace{0.12cm}

\noindent
\textbf{Visualizations of the localization comparison.} 
Figure~\ref{visualization} illustrates action localization results for the unseen class \emph{BasketballDunk} on THUMOS14 and \emph{Drinking coffee} on ActivityNet v1.3. 
Compared to Ti-FAD, our method achieves more accurate temporal boundaries. 
We further visualize the similarity between text representations and video segments. 
While Ti-FAD directly adopts action labels as text representations, our approach enriches them by injecting foreground knowledge as prompt tokens. 
We can find that our method attends more effectively to action-relevant video content, validating its ability to achieve more precise cross-modal alignment and improved localization performance.

\section{Conclusion}
In this paper, we propose DFAlign, a novel framework that leverages the foreground knowledge generated by diffusion models to bridge the semantic gap between abstract action labels and complex video content for OV-TAD task.
Specifically, we first introduce the SUC module to construct conditions by integrating action-shared and action-specific semantics.
We then present the BSD module, which progressively removes background redundancy from videos via diffusion denoising under action semantic guidance, yielding discriminative foreground knowledge.
Furthermore, we propose the FPA module that injects the extracted foreground knowledge as prompt tokens into text representations, guiding model's attention toward action-relevant segments and enabling more precise cross-modal alignment.
Extensive experiments on two benchmarks demonstrate that DFAlign achieves state-of-the-art performance.

\section{Acknowledgment}
This work was supported by the National Key R\&D Program of China under Grant 2022YFB3103503, the Natural Science Foundation of China under Grant 62576310, the Zhejiang Provincial Natural Science Foundation of China under Grant No. LQN26F020052.

\bibliographystyle{ACM-Reference-Format}
\bibliography{main}

\appendix

\end{document}